\documentclass[letterpaper, 10 pt, conference]{ieeeconf}  % Comment this line out if you need a4paper

\IEEEoverridecommandlockouts                              % This command is only needed if 
\usepackage{decar-common}
\usepackage{decar-dynamics}
\usepackage{decar-lie}
\usepackage{decar-post}
\usepackage{custom-commands}

\usepackage{booktabs}
\usepackage{stfloats}
\usepackage[backend=bibtex,bibstyle=ieee,citestyle=numeric-comp,doi=false,isbn=false,url=false]{biblatex}

\DeclareSIUnit{\sqrthz}{\ensuremath{\sqrt{\mathrm{Hz}}}}
\DeclareSIUnit{\invsqrthz}{\ensuremath{/\sqrt{\mathrm{Hz}}}}
\newif\ifanonymous
\anonymousfalse

\ifanonymous
  \author{Author Names Hidden}
\else
\author{Mitchell Cohen$^{1}$, Vassili Korotkine$^{1}$, and James Richard Forbes$^{1}$% <-this % stops a space
\thanks{*This work was supported by the Natural Sciences and Engineering Research Council of Canada (NSERC) Alliance Program and the Canadian Department of National Defence.}% <-this % stops a space
\thanks{$^{1}$The authors are with the Department of Mechanical Engineering, McGill University, Montreal, QC H3A 0C3, Canada.
        {\tt\small mitchell.cohen3@mail.mcgill.ca}%
}
}
\fi

\title{\LARGE \bf Observability and Consistency Analysis for Visual-Inertial Navigation with Anchored Feature Parameterizations}

\begin{document}

%auto-ignore
% This is not a standalone latex document. To use this file
% as a cover page on an arXiv upload of a document that is 
% already accepted as some sort of IEEE publication, you must
%
%  1) add the following just after the \begin{document} line
%     of your main paper document
%
%         \input{arxiv-cover-ieee.tex}
%
%  2) and replace the relevant information in the block below.
%
% The relevant information has been parameterized as variables.
% Simply replace the variable values with your stuff and the 
% result should be good.
%
% Make sure to not include this file for ACTUAL submissions to 
% the IEEE. Luckily you can just comment in/out the 
% \input{arxiv-cover-ieee.tex} line.
%
% FYI: The exact citation with formatting can be obtained 
% from your paper's page on IEEE Xplore.
%
%%%%%%%%%%%%%%%%%%%%%%%%%%%%%%%%%%%%%%%%%%%%%%%%%%%%%%%%%%%%%%%
%%%%%%%%%%%%%%%%%%%%%% ADD YOUR INFO HERE %%%%%%%%%%%%%%%%%%%%%
%%%%%%%%%%%%%%%%%%%%%%%%%%%%%%%%%%%%%%%%%%%%%%%%%%%%%%%%%%%%%%%
\def \myJournal {IEEE/RSJ International Conference on Intelligent Robots and Systems}
\def \myDoi {10.1109/LRA.2021.3067640}
\def \myPaperSiteName {IEEE Xplore}
\def \myPaperSiteLink {https://ieeexplore.ieee.org/document/9382085}
\def \myYear {2026}
\def \myPaperCitation{C. C. Cossette, M. Shalaby, D. Saussié, J. R. Forbes and J. Le Ny, ``Relative Position Estimation Between Two UWB Devices With IMUs,'' in \textit{IEEE Robotics and Automation Letters}, vol. 6, no. 3, pp. 4313-4320, July 2021.}

%%%%%%%%%%%%%%%%%%%%%%%%%%%%%%%%%%%%%%%%%%%%%%%%%%%%%%%%%%%%%%%
%%%%%%%%%%%%%%%%%%%%%%%%%%%%%%%%%%%%%%%%%%%%%%%%%%%%%%%%%%%%%%%

\begin{figure*}[t]

\thispagestyle{empty}
\begin{center}
\begin{minipage}{6in}
\centering
This paper has been accepted for publication at the \emph{\myJournal}. 
\vspace{1em}

This is the author's version of an article that has, or will be, published in this journal or conference. Changes were, or will be, made to this version by the publisher prior to publication.
\vspace{2em}

% \begin{tabular}{rl}
% DOI: & \myDoi\\
% \myPaperSiteName: & \texttt{\myPaperSiteLink}
% \end{tabular}

% \vspace{2em}
% Please cite this paper as:

% \myPaperCitation

\vspace{15cm}
\copyright \myYear \hspace{4pt}IEEE. Personal use of this material is permitted. Permission from IEEE must be obtained for all other uses, in any current or future media, including reprinting/republishing this material for advertising or promotional purposes, creating new collective works, for resale or redistribution to servers or lists, or reuse of any copyrighted component of this work in other works.

\end{minipage}
\end{center}
\end{figure*}
\newpage
\clearpage
\pagenumbering{arabic}

\maketitle
\thispagestyle{empty}
\pagestyle{empty}

%%%%%%%%%%%%%%%%%%%%%%%%%%%%%%%%%%%%%%%%%%%%%%%%%%%%%%%%%%%%%%%%%%%%%%%%%%%%%%%%
\begin{abstract}
        This paper presents an analysis of the observability and consistency properties of filtering-based visual-inertial navigation systems (VINS) that utilize anchored feature representations. The unobservable subspace of VINS with anchored landmark parameterizations is shown to be independent of the estimated landmark state, which leads to improved estimator consistency properties without any additional modifications. However, the unobservable subspace is still found to depend on the estimated navigation state, necessitating additional consistency-enforcing techniques. Two methods to improve the consistency of VINS with anchored feature representations are presented. Simulation results showcase that all estimators employing anchored feature paramterizations exhibit improved consistency properties compared to algorithms that estimate features resolved in a global reference frame, especially in scenarios where feature initialization may be poor. Real-world experiments on the TUM-VI dataset showcase that the use of anchored feature representations alone can yield comparable performance to consistency-improved estimators employing a global feature representation, demonstrating the benefit of using anchored feature parameterizations for VINS.
\end{abstract}

\section{Introduction}
Estimating the state of a system from noisy sensor measurements is a
critical task in many applications. In GPS-denied environments, the fusion of
cameras and inertial measurement units (IMUs) has gained significant attention in recent years and has been successfully applied to several domains, including augmented and virtual
reality and navigation of mobile robotic platforms such as self-driving cars. Visual-inertial navigation systems (VINS)
enable low cost, lightweight, and robust state estimation due to the
complementary nature of the two sensor modalities. Real-time VINS approaches
are commonly divided into two categories: filter-based
\cite{Mourikis_Roumeliotis_2007}, \cite{Geneva_Eckenhoff_Lee_Yang_Huang_2020}
and sliding-window optimization-based~\cite{Huang_Mourikis_Roumeliotis_2011},~\cite{qin2018vins},~\cite{Leutenegger_Lynen_Bosse_Siegwart_Furgale_2015}. Filtering methods, such as the multi-state constraint Kalman filter (MSCKF)~\cite{Mourikis_Roumeliotis_2007}, are based on the Extended Kalman Filter (EKF), which linearizes the process and measurement models a single time at every timestep. Optimization-based approaches iteratively relinearize process and measurement models over a window of measurements, leading to improved performance in situations with poor state initialization~\cite{chen2025iteration}. Nonetheless, filtering-based methods are still widely used for resource-constrained applications due to their computational efficiency.

\subsection{Related Work}
\vspace{-1mm}
\emph{Consistency} refers to an estimator's ability to accurately characterize the uncertainty of the estimation error, and is crucial for any VINS estimator. Ensuring consistent estimation has not only been shown to improve accuracy for VINS~\cite{chen2022fej2}, but
additionally can aid downstream tasks including motion
planning~\cite{prentice2009belief} and inferring uncertain data
associations~\cite{doherty2019multimodal}. Due to the local nature of inertial
and visual measurements, VINS have four unobservable directions,
corresponding to the global position and the rotation of the platform about the
gravity vector \cite{hesch2013consistency}. Standard EKF-based VINS estimators are known to be inconsistent and produce
overconfident state estimates, caused by mismatches in linearization points for the system Jacobians~\cite{huang2009first}. There has been significant research
dedicated to ensuring consistency of VINS estimators, and main approaches include i) first-estimate Jacobian (FEJ) techniques
\cite{huang2009first},~\cite{li2013high} ii) observability-constrained (OC)
methods~\cite{hesch2013consistency}, iii) right-invariant filtering formulations (RI-EKF)~\cite{Yang_Chen_Lee_Huang_2022},~\cite{wu2017invariant}, and iv) robocentric estimators~\cite{huai2022robocentric}. More recently, the transformation-based EKF~\cite{tian2025t} and affine EKF~\cite{song2026affine} have also been proposed to improve estimator consistency.

While most existing research on consistent visual-inertial estimation has focused on analyzing the consistency properties of algorithms that represent landmarks in a global reference frame, there exist several other choices of landmark parameterizations. The representation of the estimated environmental map is a key design choice in VINS estimators and has a significant impact on the resultant estimator performance. In~\cite{montiel2006unified} and~\cite{Civera_Davison_Montiel_2008}, an \emph{inverse-depth} representation of landmarks is presented, which represents landmarks utilizing their inverse depth relative to the camera position at which they were first observed. This parameterization is shown to offer benefits when used in visual SLAM systems compared to the standard Euclidean parameterization, especially for landmarks that are distant from the camera. A comparison between various point landmark representations is performed in~\cite{sola2012impact}, but the impact of landmark representation on observability and consistency properties of the system is not analyzed. The MSCKF-based visual-inertial estimator OpenVINS~\cite{Geneva_Eckenhoff_Lee_Yang_Huang_2020} supports several anchored and global landmark representations, where \emph{anchored} feature representations represent landmarks relative to an anchoring camera pose rather than in a global frame of reference. Several choices of anchored landmarks include the form of inverse-depth landmarks presented in~\cite{Mourikis_Roumeliotis_2007}, as well as direct Euclidean anchored landmarks as presented in~\cite{paul2017comparative}. Despite the prevalent use of anchored features in VINS, existing work has not thoroughly explored the impact of this decision on the observability and consistency properties of the resultant estimators. In particular, the influence of consistency-improving design choices, such as the use of FEJ and invariant error definitions, has not been systematically analyzed for VINS employing anchored features.

\vspace{-2mm}
\subsection{Contributions}
The main contribution of this paper is an explicit mathematical analysis, followed by simulation and experimental validation, of consistency properties of filtering-based VINS estimators that utilize anchored feature representations. Specifically, the contributions of this paper are:
\begin{itemize}
    \item A novel analysis of the unobservable subspace of VINS estimators employing anchored feature representations, revealing that the estimator unobservable subspace is independent of the landmark states, but still depends on the navigation state. This observation shows that linearization point changes of the landmark states do not impact estimator consistency.
    \item Practical design insights derived from the analysis on the application of existing consistency-improving techniques, namely FEJ and invariant error definitions, to VINS estimators using anchored features.
    % \item An analysis of the application of consistency-improving techniques, namely FEJ and invariant error definitions, to VINS estimators using anchored features.
    \item Thorough Monte-Carlo simulations with three simulated trajectories, showcasing that estimators utilizing anchored features exhibit improved consistency properties compared to estimators using global feature representations. Additional evaluations are performed on five sequences of the TUM-VI dataset~\cite{schubert2018tum} to validate the performance benefits of anchored feature representations in real-world scenarios.
\end{itemize}

% While the unobservable subspace of VINS with landmarks resolved in the global frame explicitly depends on the landmarks, it is shown that the unobservable subspace of VINS with anchored feature representations is \emph{independent} of the landmark state. This leads to improved consistency properties of estimators employing anchored feature representations, even without additional modifications. The impact of consistency-improving techniques, namely FEJ and RI-EKF, is analyzed for estimators using anchored feature representations, and compared to the estimators that utilize global feature representations through rigorous Monte-Carlo simulations, and on real-world data from the TUM-VI dataset~\cite{schubert2018tum}. Anchored feature representations are shown to offer improved consistency compared to global feature representations, and the use of consistency-improving techniques is shown to further improve estimator performance.

The rest of the paper is structured as follows. Section~\ref{sec:visual_inertial_navigation_systems} provides an overview of the visual-inertial navigation problem. Section~\ref{sec:observability_analysis} presents an observability analysis for VINS estimators utilizing both global and anchored feature representations, and Section~\ref{sec:consistency_improvements} details methods of improving the consistency of VINS algorithms using anchored features. Sections~\ref{sec:monte_carlo_simulations} and~\ref{sec:real_world_experiments} present Monte-Carlo simulations and real-world experiments comparing the performance of estimators utilizing global and anchored feature representations, as well as different consistency-improving techniques for both landmark representations. Finally, Section~\ref{sec:conclusion} provides concluding remarks and discusses future research directions.

\vspace{-1mm}
\section{Visual-Inertial Navigation Systems}
\label{sec:visual_inertial_navigation_systems}
This section will review the standard formulation of sliding-window, EKF-based VINS. The filter state vector includes the current IMU state, $\mc{X}_k^\mc{I}$, a sliding window of $c$ previous IMU pose cloned states~\cite{roumeliotis2002stochastic}, denoted $\mc{X}_k^\mc{W}$, and a set of $m$ long-tracked landmarks, denoted $\mc{X}^\mc{F}$. The state vector at time $t = t_k$ is given by
\begin{align}
    \label{eq:vins_state}
    \mc{X}_k        & = \left(\mc{X}_k^\mc{I}, \mc{X}_k^\mc{W}, \mc{X}^\mc{F} \right),                       \\
    \mc{X}_k^\mc{I} & = \left(\mbf{C}_{ab_k}, \mbf{v}_a^{z_k w/a}, \mbf{r}_a^{z_k w}, \mbf{b}_{b_k} \right), \\
    \mc{X}_k^\mc{W} & = \left(\mc{X}_{k-1}^\text{p}, \ldots, \mc{X}_{k-c}^\text{p} \right),                  \\
    \mc{X}^\mc{F}   & = \left(\mbs{\lambda}^1, \ldots, \mbs{\lambda}^m \right),
\end{align}
where $\mbf{C}_{ab_k} \in SO(3)$ is the direction-cosine-matrix at time $t = t_k$ that relates the orientation of the IMU frame $\mc{F}_{b_k}$ to the orientation of a gravity-aligned global frame $\mc{F}_a$, $\mbf{v}_a^{z_k w/a}$ and $\mbf{r}_a^{z_k w}$ are the velocity and position of the IMU resolved in $\mc{F}_a$ at time $t = t_k$, and $\mbf{b}_{b_k} = \begin{bmatrix} \mbf{b}_{b_k}^{\text{g}^\trans} & \mbf{b}_{b_k}^{\text{a}^\trans} \end{bmatrix}^\trans$ are the IMU biases at time $t = t_k$, with $\mbf{b}_{b_k}^\text{g}$ and $\mbf{b}_{b_k}^\text{a}$ being the gyroscope and accelerometer biases, respectively, resolved in $\mc{F}_{b_k}$. Additionally, the cloned pose state at time $t = t_i$ includes the attitude and position of the IMU, such that $\mc{X}_{i}^p = \left(\mbf{C}_{ab_i}, \mbf{r}_a^{z_i w} \right)$. Finally, the landmark states $\mbs{\lambda}^j, j = 1, \ldots, m$ can be represented in a variety of ways, as discussed in the next section. A given landmark $p_j$ resolved in the global frame $\mc{F}_a$ is denoted $\mbf{r}_a^{p_j w}$, and is related to the landmark state $\mbs{\lambda}^j$ through a function $\mbf{h}_r \left(\mbs{\lambda}^j, \cdot \right)$, such that $\mbf{r}_a^{p_j w} = \mbf{h}_r \left(\mbs{\lambda}^j, \cdot \right)$, which may depend on additional components of the state for anchored features, such as the anchoring pose.

An EKF-based estimator relies on a linearization of the process and measurement models, which requires defining an error on the filter state. Given a perturbation $\delta \mbs{\xi} = \begin{bmatrix}\dxi^{\mc{I}^\trans} & \dxi^{\mc{W}^\trans}& \dxi^{\mc{F}^\trans}\end{bmatrix}^\trans$, with
$\dxi^{\mc{I}} \in \mathbb{R}^{15}$, $\dxi^{\mc{W}} \in \mathbb{R}^{6c}$, and $\dxi^{\mc{F}} \in \mathbb{R}^{3m}$,
the relationship between the true state $\mc{X}_k$ and the estimated state $\mchat{X}_k$ is given by $\mc{X}_k = \hat{\mc{X}}_k \oplus \delta \mbs{\xi}$
where the $\oplus : G \times \mathbb{R}^n \to G$ operator allows for incrementing the state, defined on a manifold $G$, with a perturbation defined in the tangent space $\mathbb{R}^n$~\cite{Solà_Deray_Atchuthan_2021}. In standard VINS estimators, the $\oplus$ operator is defined such that the error on positions, velocities, biases, and landmarks are defined utilizing an additive error on a vector space, that is, $\mbf{x} = \mbfhat{x} + \delta \mbf{x}$, where $\mbf{x}$, $\mbfhat{x}$ and $\delta \mbf{x}$ are the true, estimated, and error states, respectively, all elements of $\mathbb{R}^3$. The orientation error is defined using a right perturbation on $SO(3)$, such that the relationship between a true orientation $\mbf{C}$ and the estimated orientation $\mbfhat{C}$ is given by $\mbf{C} = \mbfhat{C} \Exp \left(\dxi^\phi \right)$, where $\mbf{C} \in SO(3)$, $\dxi^\phi \in \mathbb{R}^3$, and $\Exp \left(\dxi^\phi \right) \triangleq \exp \left(\dxi^{\phi^\times} \right)$. Additionally, $\exp : \mathfrak{so}(3) \to SO(3)$ is the exponential map on $SO(3)$, and $\left(\cdot \right)^\times : \mathbb{R}^3 \to \mathfrak{so}(3)$ is the skew-symmetric operator, such that $\mbf{a}^\times \mbf{b} = -\mbf{b}^\times \mbf{a} \hspace{2mm} \forall \mbf{a}, \mbf{b} \in \mathbb{R}^3$.

In visual-inertial estimation, gyroscope and accelerometer measurements from the IMU, denoted $\mbf{u}_k = \begin{bmatrix} \mbf{u}_{b_k}^{\text{g}^\trans} & \mbf{u}_{b_k}^{\text{a}^\trans} \end{bmatrix}^\trans$, are utilized to propagate the estimated inertial state forward from time $t = t_k$ to time $t = t_{k+1}$, by discretizing the continuous-time process model given by
\begin{align}
    \label{eq:continuous_time_imu_kinematics}
    \dot{\mc{X}}^{\mc{I}} \left( t \right) & = f \left(\mc{X}^\mc{I} \left(t \right), \mbf{u} \left(t \right), \mbf{w} \left(t \right) \right),
\end{align}
where the process model can be written component-wise, with the time argument dropped for clarity, as
\begin{align}
    \mbfdot{C}_{ab}       & = \mbf{C}_{ab} \left(\mbf{u}_b^\text{g} - \mbf{b}_b^\text{g} - \mbf{w}_b^\text{g} \right)^\times,      \\
    \mbfdot{v}_a^{zw/a}   & = \mbf{C}_{ab} \left(\mbf{u}_b^\text{a} - \mbf{b}_b^\text{a} - \mbf{w}_b^\text{a} \right) + \mbf{g}_a, \\
    \label{eq:continuous_time_imu_kinematics_last}
    \mbfdot{r}_a^{zw}     &= \mbf{v}_a^{zw/a}, \hspace{4mm}
    \mbfdot{b}_b^\text{g} = \mbf{w}_b^{b_g}, \hspace{4mm}
    \mbfdot{b}_b^\text{a} = \mbf{w}_b^{b_a},
\end{align}
where $\mbf{g}_a = \begin{bmatrix} 0 & 0 & -g\end{bmatrix}^\trans$ is the gravity vector resolved in $\mc{F}_a$ with the gravity magnitude $g$, $\mbf{w}_b^g \sim \mc{N} \left(\mbf{0}, \mbs{\mc{Q}}^g \delta \left(t - \tau \right) \right)$, $\mbf{w}_b^a \sim \mc{N} \left(\mbf{0}, \mbs{\mc{Q}}^a \delta \left(t - \tau \right) \right)$ are white noises on the gyroscope and accelerometer measurements respectively, with power spectral densities $\mbs{\mc{Q}}^g$ and $\mbs{\mc{Q}}^a$, and $\delta \left(\cdot \right)$ is the Dirac delta function. The IMU biases are modelled as random walk processes with $\mbf{w}_b^{b_g} \sim \mc{N} \left(\mbf{0}, \mbs{\mc{Q}}^{b_g} \delta (t -  \tau) \right)$ and $\mbf{w}_b^{b_a} \sim \mc{N} \left(\mbf{0}, \mbs{\mc{Q}}^{b_a} \delta \left(t - \tau \right) \right)$. Defining the overall process noise as $\mbf{w} = \begin{bmatrix} \mbf{w}_b^{g^\trans} & \mbf{w}_b^{a^\trans} & \mbf{w}_b^{b_g^\trans} & \mbf{w}_b^{b_a^\trans} \end{bmatrix}^\trans$, with $\mbf{w} \sim \mathcal{N} \left(\mbf{0}, \mc{\mbs{\mc{Q}}}_c \delta (t - \tau) \right)$ the linearized continuous-time dynamics of the error for the entire filter state are given by
\begin{align}
    \delta \mbsdot{\xi} & =
    \underbrace{\begin{bmatrix}
            \mbf{F}^\mc{I} (t) & \mbf{0} & \mbf{0} \\
            \mbf{0}            & \mbf{1} & \mbf{0} \\
            \mbf{0}            & \mbf{0} & \mbf{1}
        \end{bmatrix}}_{\mbf{F} \left(t \right)}
    \underbrace{\begin{bmatrix}
            \dxi^\mc{I} \\
            \dxi^\mc{W} \\
            \dxi^\mc{F}
        \end{bmatrix}}_{\dxi} +
    \underbrace{\begin{bmatrix}
            \mbf{L}^\mc{I} (t) \\
            \mbf{0}            \\
            \mbf{0}
        \end{bmatrix}}_{\mbf{L} (t)}
    \mbf{w} (t),
\end{align}
where $\mbf{F}^\mc{I} \left(t \right)$ is the Jacobian of the continuous-time IMU process model with respect to $\mc{X}^\mc{I}$ and $\mbf{L}^\mc{I} (t)$ is the Jacobian of the process model with respect to the noise. The state covariance propagation relies on the discrete-time system state-transition matrix $\mbs{\Phi}_{k+1, k}$ where $\mbs{\Phi}_{k+1, k}$ is the solution to the differential equation~\cite{farrell2008aided}
\begin{align}
    \mbsdot{\Phi}_{k+1, k} = \mbf{F} \left(t \right) \mbs{\Phi}_{k+1, k}, \hspace{5mm} \mbs{\Phi}_{k, k} = \mbf{1}.
\end{align}
The expression for $\mbs{\Phi}_{k+1, k}$ can either be found numerically or analytically~\cite{hesch2013consistency}, and can be used to propagate the filter state covariance as $\mbfcheck{P}_{k+1} = \mbs{\Phi}_{k+1, k} \mbfhat{P}_k \mbs{\Phi}^\trans_{k+1, k} + \mbf{Q}_k$, where $\mbfhat{P}_k$ is the corrected covariance at time $t = t_k$, and $\mbfcheck{P}_{k+1}$ is the predicted covariance at time $t = t_{k+1}$. Additionally, $\mbf{Q}_k$ is found as
\begin{align}
    \mbf{Q}_k = \int_{t_k}^{t_{k+1}} \mbs{\Phi}_{k+1, \tau} \mbf{L} (\tau) \mbs{\mc{Q}}_c \mbf{L}^\trans (\tau) \mbs{\Phi}^\trans_{k+1, \tau} \mathrm{d} \tau.
\end{align}
Following a similar formulation utilized by~\cite{Geneva_Eckenhoff_Lee_Yang_Huang_2020}, visual measurements of a 3D feature $j$ detected from a given calibrated perspective camera at time $k$ can be written as a composition of functions as
\begin{align}
    \mbf{y}_{jk}                                    & = \mbf{g} \left(\mc{X}_k \right) + \mbf{v}_{jk},                                                                               \\
    \mbf{g} \left(\mc{X}_k \right)                  & = \mbf{h}_p \left(\mbf{r}_{c_k}^{p_j c_k} \right),                                                                             \\
    \mbf{h}_p \left(\mbf{r}_{c_k}^{p_j c_k} \right) & =
    \begin{bmatrix}
        x_k^j /z_k^j &
        y_k^j/z_k^j
    \end{bmatrix}^\trans, \hspace{2mm}
    \mbf{r}_{c_k}^{p_j c_k} = \begin{bmatrix} x_k^j & y_k^j & z_k^j \end{bmatrix}^\trans,                                                                                                                     \\
    \mbf{r}_{c_k}^{p_j c_k}                         & = \mbf{C}_{bc}^\trans \left(\mbf{C}_{ab_k}^\trans \left(\mbf{r}_a^{p_jw} - \mbf{r}_a^{z_k w} \right) - \mbf{r}_b^{cz} \right),
\end{align}
where $\mbf{v}_{jk} \sim \mathcal{N} \left(\mbf{0}, \mbf{R}_{jk} \right)$ and $\mbf{C}_{bc} \in SO(3)$, and $\mbf{r}_b^{cz} \in \mathbb{R}^3$ are the extrinsic parameters of the system, and $\mbf{r}_a^{p_j w} = \mbf{h}_r \left(\mbs{\lambda}^j, \cdot \right)$. The measurement model Jacobian has the form
\begin{align}
    \label{eq:measurement_jacobian}
    \mbf{H}_k         & = \mbf{H}_k^\pi \mbf{C}_{ab_k}^\trans
    \begin{bmatrix}
        \mbf{H}_k^\phi & \mbf{0} & -\mbf{1} & \mbf{0} & \mbf{0} & \vline & \mbf{H}_k^p & \vline & \mbf{H}_k^\lambda
    \end{bmatrix},                                                                                \\
    \mbf{H}_k^\pi     & = \frac{1}{z_k^j} \begin{bmatrix}
        1 & 0 & -x_k^j/z_k^j \\
        0 & 1 & -y_k^j/z_k^j
    \end{bmatrix} \mbf{C}_{bc}^\trans,                      \\
    \mbf{H}_k^\phi    & = \left( \mbfbar{r}_a^{p_j w} - \mbfbar{r}_a^{z_k w} \right)^\times \mbfbar{C}_{ab_k}, \\
    \mbf{H}_k^p       & =
    \begin{bmatrix}
        \mbf{0} & \cdots & \mbf{H}_k^{p_i} & \cdots & \mbf{0}
    \end{bmatrix},                                                                                \\
    \mbf{H}_k^\lambda & =
    \begin{bmatrix}
        \mbf{0} & \cdots & \mbf{H}_k^{\lambda_j} & \cdots & \mbf{0}
    \end{bmatrix},
\end{align}
where $\mbf{H}_k^{p_i}$ is a Jacobian component of the measurement model with respect to the anchoring pose $p_i$, and $\mbf{H}_k^\lambda$ is the Jacobian component with respect to the landmark states, whose expressions both depend on the specific landmark representation utilized.

\subsection{Landmark Representations}
This work focuses on two main landmark representations. The first representation is a Euclidean 3D parameterization, where the three components of the landmark's position are estimated as directly resolved in the global frame, such that $\mbs{\lambda}^j = \mbf{r}_a^{p_j w}$ and $\mbf{H}_k^{p_i} = \mbf{0}$, $\mbf{H}_k^{\lambda_j} = \mbf{1}$. Anchored parameterizations estimate a nonlinear function of the position of the landmark relative to an anchoring camera frame. Denoting the anchoring camera frame as $\mc{F}_{c_f}$, the position of landmark $j$ resolved in this frame is written as $\mbf{r}_{c_f}^{p_j c_f}$. For anchored representations, the function $\mbf{h}_r \left(\mbs{\lambda}^j, \mbf{C}_{ab_f}, \mbf{r}_a^{z_f w} \right)$ additionally depends on the anchoring IMU pose and is written as
\begin{align}
    \label{eq:anchored_landmark_function}
    \mbf{r}_a^{pw} & = \mbf{h}_r \left(\mbs{\lambda}^j, \mbf{C}_{ab_f}, \mbf{r}_a^{p_f w} \right)                                              \\
                   & = \mbf{C}_{ab_f} \left(\mbf{C}_{bc} \mbf{h}_a \left(\mbs{\lambda}^j \right) + \mbf{r}_b^{cz} \right) + \mbf{r}_a^{z_f w},
\end{align}
where $\mbf{r}_{c_f}^{p_j c_f} = \mbf{h}_a \left(\mbs{\lambda}^j \right)$ is a representation-dependent nonlinear function of the landmark resolved in the anchoring frame. For any anchored representation, the Jacobian components with respect to the anchor state and landmark state in~\eqref{eq:measurement_jacobian} are given by
\begin{align}
    \label{eq:anchor_state_jacobian}
    \mbf{H}_k^{p_f} =
    \begin{bmatrix}
        -\mbfbar{C}_{ab_f} \mbfbar{r}_b^{pz^\times} & \mbf{1}
    \end{bmatrix}, \hspace{5mm} \mbf{H}_k^{\lambda_j} = \mbfbar{C}_{ab_f} \mbf{C}_{bc} \frac{\partial \mbf{h}_a}{\partial \mbs{\lambda^j}},
\end{align}
where the last partial derivative is parameterization-dependent. This paper considers the \emph{anchored inverse-depth} landmark parameterization from~\cite{Mourikis_Roumeliotis_2007}, with
\begin{align}
    \mbf{h}_a \left(\mbs{\lambda}^j \right) & = \frac{1}{\rho^j}
    \begin{bmatrix}
        \alpha^j &
        \beta^j  &
        1
    \end{bmatrix}^\trans,
\end{align}
where $\mbs{\lambda}^j = \begin{bmatrix} \alpha^j & \beta^j & \rho^j \end{bmatrix}^\trans$. When the anchoring pose $\mc{X}_f^\mathrm{p}$ is marginalized out of the sliding window, landmarks anchored to it must be reanchored to a new pose in the sliding window, requiring a mean and covariance update that incurs negligible computational overhead relative to the overall EKF update.

\vspace{-2mm}
\section{Observability Analysis}
\label{sec:observability_analysis}
The observability properties of visual-inertial estimators are critical for designing consistent state estimation algorithms, as they reveal which states can be inferred from the available measurements. Observability for EKF-based VINS estimators is typically analyzed by employing the local observability matrix~\cite{chen1990local}, defined as
\begin{align}
    \label{eq:observability_matrix}
    \mbf{M} \left(\mc{X} \right) =
    \begin{bmatrix}
        \mbf{H}_1                   \\
        \mbf{H}_2 \mbs{\Phi}_{2, 1} \\
        \vdots                      \\
        \mbf{H}_k \mbs{\Phi}_{K, 1}
    \end{bmatrix},
\end{align}
where $\mbf{H}_K$ is the measurement Jacobian at time $t_K$, and $\mbs{\Phi}_{K, 1} = \mbs{\Phi}_{K, K-1} \cdots \mbs{\Phi}_{2, 1}$ is the state transition matrix from time $t = t_1$ to $t = t_K$. The nullspace of the observability matrix, $\mbf{N}$, such that $\mbf{M} \mbf{N} = \mbf{0}$, describes the unobservable directions of the system. Consider a state vector containing the current inertial state, a single cloned pose, denoted $\mc{X}_f^\mathrm{p}$, and a single landmark $\mbs{\lambda}^1$ in the state, written as
\begin{align}
    \label{eq:analysis_state}
    \mc{X}_k = \left(\mc{X}_k^\mc{I}, \mc{X}_f^\text{p}, \mbs{\lambda}^1 \right).
\end{align}
With this state definition, it can be shown that the nullspace of the observability matrix is spanned by the columns of
\begin{align}
    \label{eq:ideal_nullspace}
    \mbf{N} =
    \begin{bmatrix}
        \mbf{0}_{3 \times 3} & \mbf{C}_{ab_1}^\trans \mbf{g}_a       \\
        \mbf{0}_{3 \times 3} & -\mbf{v}_a^{z_1 w/a^\times} \mbf{g}_a \\
        \mbf{1}_{3 \times 3} & -\mbf{r}_a^{z_1 w^\times} \mbf{g}_a   \\
        \mbf{0}_{3 \times 3} & \mbf{0}_{3 \times 1}                  \\
        \mbf{0}_{3 \times 3} & \mbf{0}_{3 \times 1}                  \\
        \mbf{0}_{3 \times 3} & \mbf{C}_{ab_f}^\trans \mbf{g}_a       \\
        \mbf{1}_{3 \times 3} & -\mbf{r}_a^{z_f w^\times} \mbf{g}_a   \\
        \mbf{N}_1^\lambda    & \mbf{N}_2^\lambda
    \end{bmatrix} =
    \begin{bmatrix}
        \mbf{N}^t & \vline & \mbf{N}^r
    \end{bmatrix},
\end{align}
where the first three columns $\mbf{N}^t \in \mathbb{R}^{24 \times 3}$ corresponds to global position and the last column $\mbf{N}^r \in \mathbb{R}^{24 \times 1}$ corresponds to rotation about the gravity vector. Additionally, $\mbf{N}_1^\lambda$ and $\mbf{N}_2^\lambda$ are the nullspace components related to the landmark states, whose expressions depend on the representation used for the landmarks. The standard EKF linearizes process and measurement models about the current state, leading to change in linearization points of the same physical quantity at different times. This causes $\mbf{N}^r$ to no longer lie in the nullspace of the estimator observability matrix, causing the rotation about the gravity vector to appear observable, which can lead to inconsistent state estimates~\cite{hesch2013consistency}.
When the estimated landmark states are resolved in the global frame, such that $\mbs{\lambda}^1 = \mbf{r}_a^{p_1 w}$, the nullspace components related to the landmark are given by~\cite{hesch2013consistency}
\begin{align}
    \mbf{N}_1^\lambda & = \mbf{1}_{3 \times 3}, \hspace{5mm}
    \mbf{N}_2^\lambda = -\mbf{r}_a^{p_1 w^\times} \mbf{g}_a.
\end{align}
Notably, the nullspace depends on the landmark states, and linearization point changes in these states impacts the estimator unobservable directions. In contrast, when an anchored landmark parameterization is utilized, such that $\mbf{r}_a^{p_1 w} = \mbf{h}_r \left(\mbf{C}_{ab_f}, \mbf{r}_a^{z_f w}, \mbs{\lambda}^1 \right)$, where the form of the function $\mbf{h}_r \left(\cdot \right)$ is given by~\eqref{eq:anchored_landmark_function}, the nullspace components corresponding to the landmark states are given by
\begin{align}
    \mbf{N}_1^\lambda & = \mbf{0}_{3 \times 3}, \hspace{5mm}
    \mbf{N}_2^\lambda = \mbf{0}_{3 \times 1}.
\end{align}
The proof that $\mbf{N} \mbf{M} = \mbf{0}$ with these definitions of $\mbf{N}_1^\lambda$ and $\mbf{N}_2^\lambda$ for anchored feature parameterizations in the form~\eqref{eq:anchored_landmark_function} is provided in the Appendix. Hence, the unobservable subspace is \emph{independent} of the landmark states, meaning that linearization point changes in the landmark states do not cause spurious information gain. This property of anchored landmark representations presents important implications for the design of consistent estimators, as will be discussed in the next section.

\section{Consistent Estimators Utilizing Anchored Landmark Parameterizations}
\label{sec:consistency_improvements}
This section will detail methods to ensure consistency for VINS estimators employing anchored feature parameterizations, and the differences in the application of these methods between anchored and non-anchored feature representations.

% The observability analysis presented in the previous section reveals that utilizing anchored landmark representations results in an unobservable subspace that is independent of the landmark states, meaning that their treatment does not need to be modified from the standard estimator to ensure consistency. However, dependence of the nullspace on the IMU navigation state still remains, and hence, the IMU navigation state linearization point still impacts estimator consistency. 

% \subsection{First-Estimate Jacobians (FEJ) and FEJ2}
\subsection{First-Estimate Jacobian Approach}
In the FEJ approach~\cite{huang2009first}, the process and measurement models are linearized about the \emph{first} estimate of any state that appears in the nullspace~\eqref{eq:ideal_nullspace}. When anchored feature representations are utilized, the FEJ technique \emph{only} needs to be applied to the IMU navigation state, as the nullspace is independent of the feature positions. This presents a clear benefit in situations where the initial estimate of the landmark state might be poor, such as under large noises and low-parallax feature triangulation from monocular cameras, since Jacobians can still be evaluated at the best estimate of the landmark states.
% When non-anchored feature representations are utilized, all Jacobians must be evaluated at the first estimate of the feature, which can lead to degraded performance when the initial feature estimate is poor.

% The FEJ2 approach, as presented in~\cite{chen2022fej2}, presents a modification to the standard FEJ approach, which explicitly accounts for the linearization errors incurred by the standard FEJ approach. The FEJ2 compensates for the linearization point change between the first and current state estimate, by defining the change in Jacobian a>s
%     \begin{align}
%         \Delta \mbf{H}_k = \mbfhat{H}_k - \mbfbar{H}_k,
%     \end{align}
% where $\mbfhat{H}_k$ is the Jacobian evaluated at the current state estimate and $\mbfbar{H}_k$ is the Jacobian evaluated at the first state estimate. The FEJ2 approach then defines a more accurate linear model, and projects that linear model onto the left nullspace of $\Delta \mbf{H}$, which 

% \subsection{Observability-Constrained EKF (OC-EKF)}
% Rather than hold the Jacobians fixed at the first estimate, the OC-EKF approach enforces the unobservable subspace of the estimator to span the correct unobservable directions through direct modification of the state transition matrix and measurement model Jacobians. When 

\subsection{Right-Invariant EKF (RI-EKF)}
The RI-EKF~\cite{Yang_Chen_Lee_Huang_2022},~\cite{wu2017invariant} exploits the problem structure of the underlying system by defining the filter state error in a way that renders the nullspace of the observability matrix \emph{independent} of the estimated state, rendering the unobservable subspace invariant to linearization point changes. For VINS estimators with features resolved in the global frame, the RI-EKF couples the error definition of the landmark to the error definition of the IMU navigation state, leading to increased computational cost in the EKF prediction step~\cite{Yang_Chen_Lee_Huang_2022}. Hence,~\cite{Yang_Chen_Lee_Huang_2022} proposes an estimation scheme that decouples the IMU navigation state error from the landmark state, requiring FEJ to only be used on the feature position. This scheme, termed DRI-FEJ, still has the disadvantage of linearizing about the first estimate of the landmark state.

When the landmarks are estimated in an anchoring frame, only the IMU navigation state error and cloned pose error need to be defined utilizing a right-invariant error definition to ensure consistency. The right-invariant perturbation on the IMU state is defined as
\begin{align}
    \mc{\bar{X}}^\mc{I} \oplus \dxi^\mc{I} =
    \begin{pmatrix}
        \Exp \left(\dxi^\phi \right) \mbfbar{C}_{ab}                                                    \\
        \Exp \left(\dxi^\phi \right) \mbfbar{v} + \mbf{J}^\ell \left(\dxi^\phi \right) \dxi^\mathrm{v}  \\
        \Exp \left(\dxi^\phi \right) \mbfbar{r} + \mbf{J}^\ell \left( \dxi^\phi \right) \dxi^\mathrm{r} \\
        \mbfbar{b}_b + \dxi^b
    \end{pmatrix},
\end{align}
where $\mbf{J}^\ell \left(\cdot \right)$ is the left Jacobian on $SO(3)$, and $\dxi^\mc{I} = \begin{bmatrix} \dxi^{\phi^\trans} & \dxi^{\mathrm{v}} & \dxi^{\mathrm{r}^\trans} & \dxi^{b^\trans} \end{bmatrix}^\trans \in \mathbb{R}^{15}$ represents the perturbation applied to the IMU state. Perturbations to the cloned IMU pose are defined in an analogous fashion. Utilizing these perturbations yields a different form of the discrete-time IMU state transition matrix and measurement model Jacobians, where the exact expressions of these can be found in~\cite{Yang_Chen_Lee_Huang_2022}. When utilizing the RI-EKF with anchored feature representations, the unobservable subspace is \emph{state-estimate independent}, meaning the RI-EKF with anchored features is theoretically  consistent without further modifications.
\vspace{-1mm}
\section{Monte-Carlo Simulations}
\label{sec:monte_carlo_simulations}
Monte-Carlo simulations are performed to analyze the impact of the two presented consistency-improved estimator designs for anchored landmark parameterizations. The visual-inertial simulator included in~\cite{Geneva_Eckenhoff_Lee_Yang_Huang_2020} is utilized to generate inertial measurements and camera pixel readings from landmarks along three simulated trajectories. The first two trajectories, the TUM Corridor  and Udel Gore trajectories, are similarly utilized in~\cite{Yang_Chen_Lee_Huang_2022}, and have respective lengths of \SI{227}{\meter} and \SI{290}{\meter}. The third trajectory, Udel ARL, is a 30 minute, 2.4-\textrm{km} trajectory as utilized in~\cite{peng2025sqrt}. The configuration used for all simulations is shown in Table~\ref{tab:monte_carlo_config}.
\begin{table}[t]
    {\setlength{\tabcolsep}{4pt}
        \footnotesize
        \centering
        \caption{Monte-Carlo Simulation Parameters}
        \label{tab:monte_carlo_config}
        \vspace{-4pt}
        \resizebox{\columnwidth}{!}{%
            \begin{tabular}{cccc}
                \hline
                \textbf{Parameter}   & \textbf{Value}                                   & \textbf{Parameter} & \textbf{Value}                                    \\
                \hline
                Accel. White Noise   & \SI{2.0e-3}{\meter\per\second\squared\invsqrthz} & Gyro. White Noise  & \SI{1.7e-04}{\radian\per\second\invsqrthz}        \\
                Accel. Random Walk   & \SI{3.0e-3}{\meter\per\second\cubed\invsqrthz}   & Gyro. Random Walk  & \SI{1.9e-5}{\radian\per\second\squared\invsqrthz} \\
                IMU Freq.            & \SI{400}{\hertz}                                 & Cam Freq.          & \SI{10}{\hertz}                                   \\
                Max Cam Points/Frame & 100                                              & Number of Cams     & 2                                                 \\
                \hline
            \end{tabular}}
    }
\end{table}

OpenVINS~\cite{Geneva_Eckenhoff_Lee_Yang_Huang_2020} is used as the base estimator for all experiments, which already supports the use of global, anchored Euclidean and anchored inverse-depth landmark representations. Two feature representations are compared, those being the baseline global Euclidean 3D parameterization (G3D), and the anchored inverse-depth parameterization (AID) described in Section~\ref{sec:visual_inertial_navigation_systems}. Anchored Euclidean landmarks were also evaluated and were found to produce similar conclusions to the AID parameterization, and hence their results are omitted for brevity. For each feature representation, three estimator configurations are evaluated: 1) ``Std'', the standard EKF with no consistency improvements; 2) the EKF utilizing FEJ; and 3) the RI-EKF, denoted RI. While OpenVINS~\cite{Geneva_Eckenhoff_Lee_Yang_Huang_2020} already supports the use of FEJ with these landmark representations, the RI-EKF is integrated into OpenVINS utilizing the expressions for the discrete-time state-transition matrix and measurement Jacobians found in~\cite{Yang_Chen_Lee_Huang_2022}. For the RI-G3D configuration, the decoupled right-invariant scheme described in~\cite{Yang_Chen_Lee_Huang_2022} is used, where FEJ is only applied to the landmark states, and the best estimate of the IMU navigation state is used in computing Jacobians. For the RI-AID configuration, FEJ is not required to ensure estimator consistency, and all Jacobians are evaluated at the best estimate of the state, as described in Section~\ref{sec:consistency_improvements}.

Following~\cite{Geneva_Eckenhoff_Lee_Yang_Huang_2020}, short-tracked features are processed within an MSCKF update and are immediately marginalized. Up to 25 long-tracked features are added into the state vector using delayed initialization~\cite{li2014visual}, and are used within the EKF update until tracking is lost. For all simulations, calibration parameters are initialized to the groundtruth and online calibration is disabled. 50 Monte-Carlo trials are run for each configuration, and the mean attitude and position RMSE at each timestep, as well as absolute trajectory error (ATE) are used to evaluate estimator accuracy~\cite{zhang2018tutorial}, while the normalized estimation error squared (NEES) is used to evaluate estimator consistency.

\begin{figure}[t]
    % \begin{figure}[htbp]
    \centering
    \includegraphics[width=0.9\columnwidth]{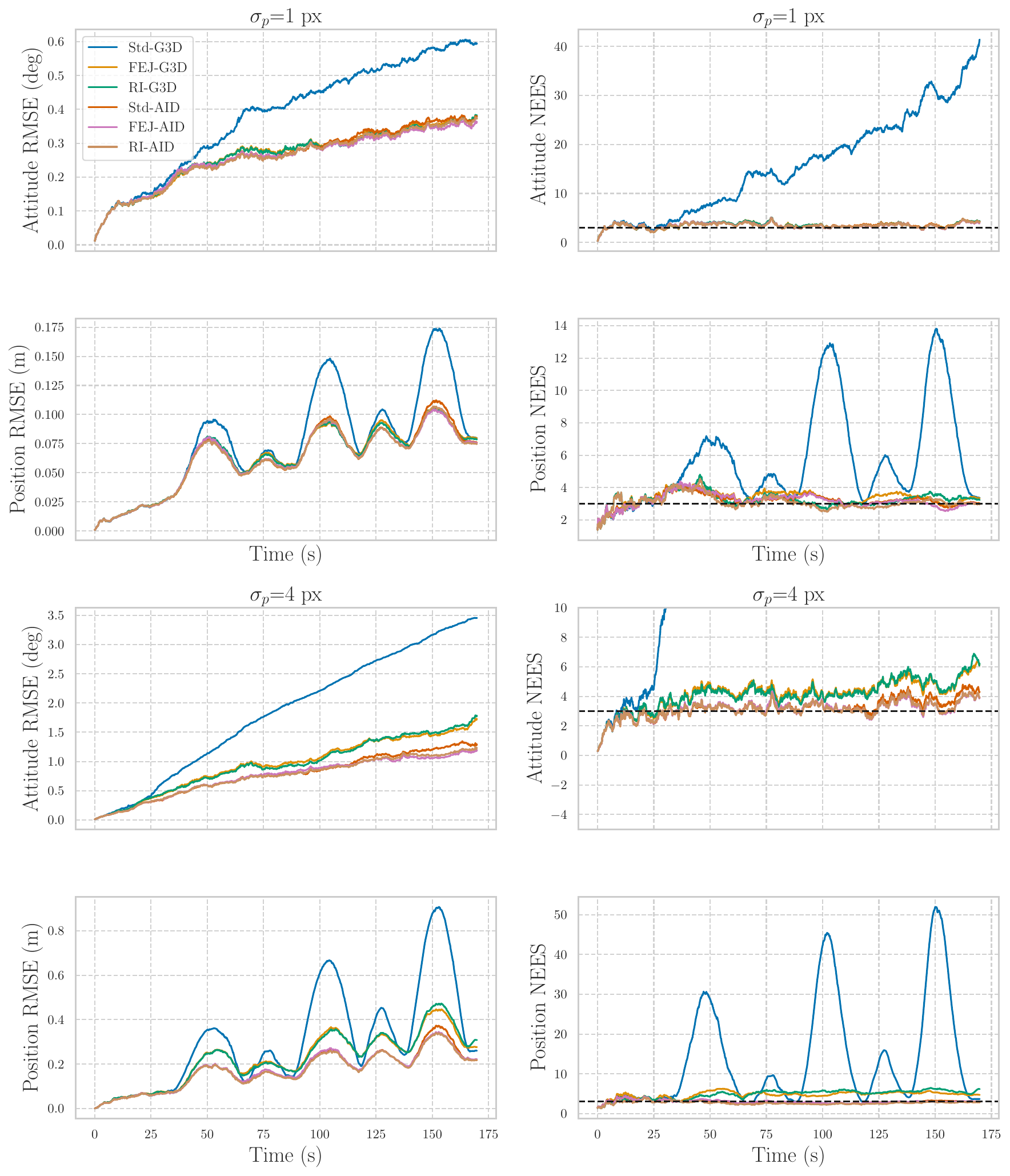}
    \vspace{-3mm}
    \caption{Orientation and position RMSEs and NEES over time averaged over 50 simulated Monte-Carlo trials for each estimator configuration for noise levels $\sigma_p = 1$ (top) and $\sigma_p = 4$ (bottom). The expected value of the NEES for a consistent estimator is represented by the dashed black line.}
    \label{fig:rmse_over_time_udel_gore}
\end{figure}

\begin{figure}[t]
    \centering
    \includegraphics[width=0.8\columnwidth]{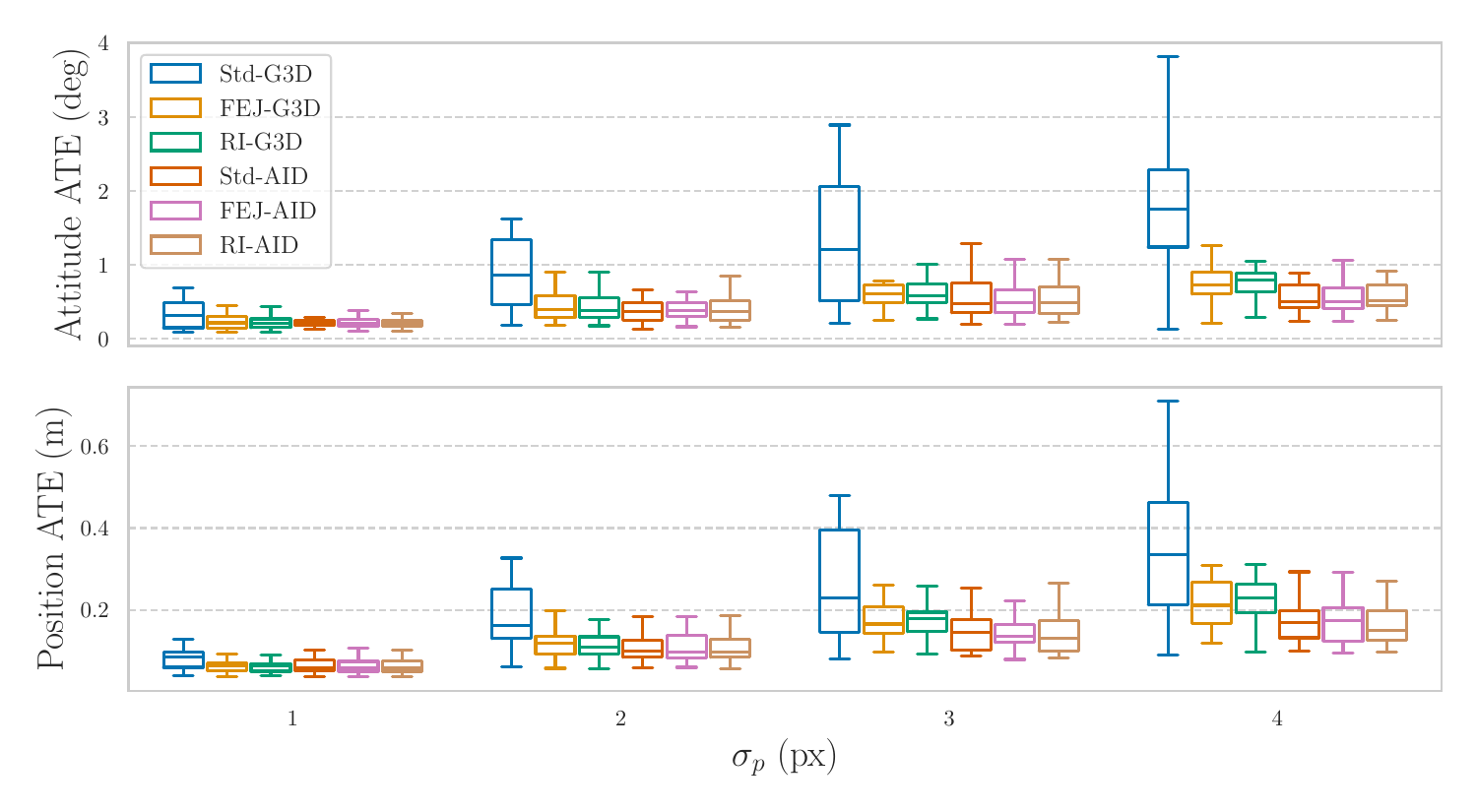}
    \vspace{-5mm}
    \caption{Distribution of position and orientation ATEs for each estimator configuration on the simulated Udel Gore trajectory, across varying camera noise levels.}
    \label{fig:ate_boxplot_udel_gore}
\end{figure}

Monte-Carlo simulations are performed for varying levels of camera noise, where the standard deviation of the pixel noise added to the simulated camera measurements is varied across $\sigma_p \in \left\{1, 2, 3, 4 \right\} \mathrm{px}$. Fig.~\ref{fig:rmse_over_time_udel_gore} and Fig.~\ref{fig:ate_boxplot_udel_gore} show the simulation results for all six estimator configurations on the Udel Gore trajectory. The mean ATEs for all estimators for all trajectories for $\sigma_p = 1 \; \mathrm{px}$ and $\sigma_p = 4 \; \mathrm{px}$ are shown in Table~\ref{tab:monte_carlo_ate}, while the mean NEES values for all estimators for all trajectories for $\sigma_p = 4 \; \mathrm{px}$ are shown in Table~\ref{tab:monte_carlo_nees}.

\begin{table}[b!]
    \centering
    \caption{Average ATEs computed over 50 Monte-Carlo trials for each estimator, on different simulated trajectories, reported as attitude ATE (degrees) / position ATE (meters).}
    \label{tab:monte_carlo_ate}
    \vspace{-2mm}
    \resizebox{\columnwidth}{!}{%
        \begin{tabular}{lcccccc}
            \toprule
                      & \multicolumn{3}{c}{$\sigma_p = 1 \; \mathrm{px}$} & \multicolumn{3}{c}{$\sigma_p = 4 \; \mathrm{px}$}                                                                                                              \\
            \cmidrule(lr){2-4} \cmidrule(lr){5-7}
            Algorithm & TUM Corridor                                      & Udel Gore                                         & Udel ARL               & TUM Corridor           & Udel Gore              & Udel ARL                        \\
            \midrule
            Std-G3D   & 0.147 / 0.066                                     & 0.356 / 0.084                                     & 1.540 / 0.170          & 0.903 / 0.310          & 1.872 / 0.351          & 15.300 / 1.115                  \\
            FEJ-G3D   & 0.129 / 0.062                                     & 0.247 / 0.066                                     & 0.882 / \textbf{0.140} & 0.486 / 0.231          & 0.878 / 0.235          & 3.901 / 0.522                   \\
            RI-G3D    & 0.130 / 0.062                                     & 0.244 / 0.066                                     & \textbf{0.868} / 0.142 & \textbf{0.395} / 0.221 & 0.901 / 0.239          & 3.868 / 0.521                   \\
            Std-AID   & 0.131 / 0.060                                     & 0.246 / 0.065                                     & 1.075 / 0.150          & 0.505 / 0.208          & 0.699 / 0.182          & 4.078 / 0.478                   \\
            FEJ-AID   & \textbf{0.127} / 0.063                            & 0.246 / 0.065                                     & 0.908 / 0.141          & 0.469 / \textbf{0.199} & \textbf{0.681} / 0.182 & \textbf{2.927} / \textbf{0.435} \\
            RI-AID    & 0.128 / \textbf{0.060}                            & \textbf{0.243} / \textbf{0.064}                   & 0.904 / 0.141          & 0.486 / 0.203          & 0.687 / \textbf{0.176} & 2.935 / 0.435                   \\
            \bottomrule
        \end{tabular}
    }
\end{table}
\begin{table}[b!]
    \centering
    \caption{Mean NEES computed over 50 simulated Monte-Carlo trials for each estimator for $\sigma_p = 4 \; \mathrm{px}$. Expected NEES values are 3 for both attitude and position.}
    \label{tab:monte_carlo_nees}
    \vspace{-1mm}
    \resizebox{\columnwidth}{!}{%
        \begin{tabular}{lcccccc}
            \toprule
                      & \multicolumn{2}{c}{TUM Corridor} & \multicolumn{2}{c}{Udel Gore} & \multicolumn{2}{c}{Udel ARL}                                     \\
            \cmidrule(lr){2-3} \cmidrule(lr){4-5} \cmidrule(lr){6-7}
            Algorithm & Att. NEES                        & Pos. NEES                     & Att. NEES                    & Pos. NEES & Att. NEES & Pos. NEES \\
            \midrule
            Std-G3D   & 21.721                           & 15.724                        & 211.475                      & 13.941    & 12093.481 & 32.145    \\
            FEJ-G3D   & 3.201                            & 5.783                         & 4.238                        & 4.825     & 4.513     & 5.376     \\
            RI-G3D    & 2.934                            & 5.886                         & 4.212                        & 4.931     & 4.394     & 5.051     \\
            Std-AID   & 3.237                            & 3.984                         & 3.284                        & 2.957     & 11.764    & 5.174     \\
            FEJ-AID   & 3.003                            & 3.415                         & 3.179                        & 3.051     & 3.557     & 4.059     \\
            RI-AID    & 3.127                            & 3.629                         & 3.164                        & 2.702     & 3.569     & 3.923     \\
            \bottomrule
        \end{tabular}
    }
\end{table}
As shown in Fig.~\ref{fig:rmse_over_time_udel_gore} and Fig.~\ref{fig:ate_boxplot_udel_gore}, all estimators employing anchored feature representations perform similarly at each noise level on the Udel Gore trajectory. This is in contrast to the estimators utilizing the global landmark representation, where the performance of Std-G3D degrades significantly as the noise level increases. While the use of consistency-improving techniques leads to more accurate performance when using a global feature representation, the accuracy of FEJ-G3D and RI-G3D still degrades at higher noise levels, since the use of FEJ on the landmark states causes the Jacobians to be more inaccurate when the initial landmark estimate is poor. In contrast, since all estimators using the AID landmark representation compute Jacobians at the \emph{best} estimate of the landmark state, their performance is less impacted by poor landmark initialization caused by high noise levels. Interestingly, the Std-AID configuration still results in near-consistent estimation for the two shorter datasets, TUM Corridor and Udel Gore, as the attitude and position NEES remain near the expected value of $3$, as shown in Table~\ref{tab:monte_carlo_nees} and the bottom rows of Fig.~\ref{fig:rmse_over_time_udel_gore}. However, the benefits of consistency-improvements for anchored feature representations can be seen for the longer Udel ARL trajectory. The information gain caused by mismatch in the linearization point of the navigation state cause the average NEES of Std-AID to be slightly higher than the expected NEES of 3 on the Udel ARL trajectory, while FEJ-AID and RI-AID more closely match expected NEES value of 3. However, the Std-AID estimator has a NEES much closer to the expected value than that of Std-G3D, highlighting that for VINS estimators employing global feature representations, inconsistency caused by information gain along unobservable directions is \textbf{primarily} caused by linearization point mismatch in the landmark states rather than mismatch in the navigation states. Anchored landmark parameterizations all have the advantage that the unobservable subspace is not impacted by the landmark state linearization point, and thus \textbf{estimators without any additional consistency improvements are more consistent than their counterparts utilizing global landmark parameterizations.}

% \subsection{Robustness to Initialization}
\section{Real-World Experiments}
\label{sec:real_world_experiments}
\begin{figure*}[t!]
    \centering
    \includegraphics[width=0.95\textwidth]{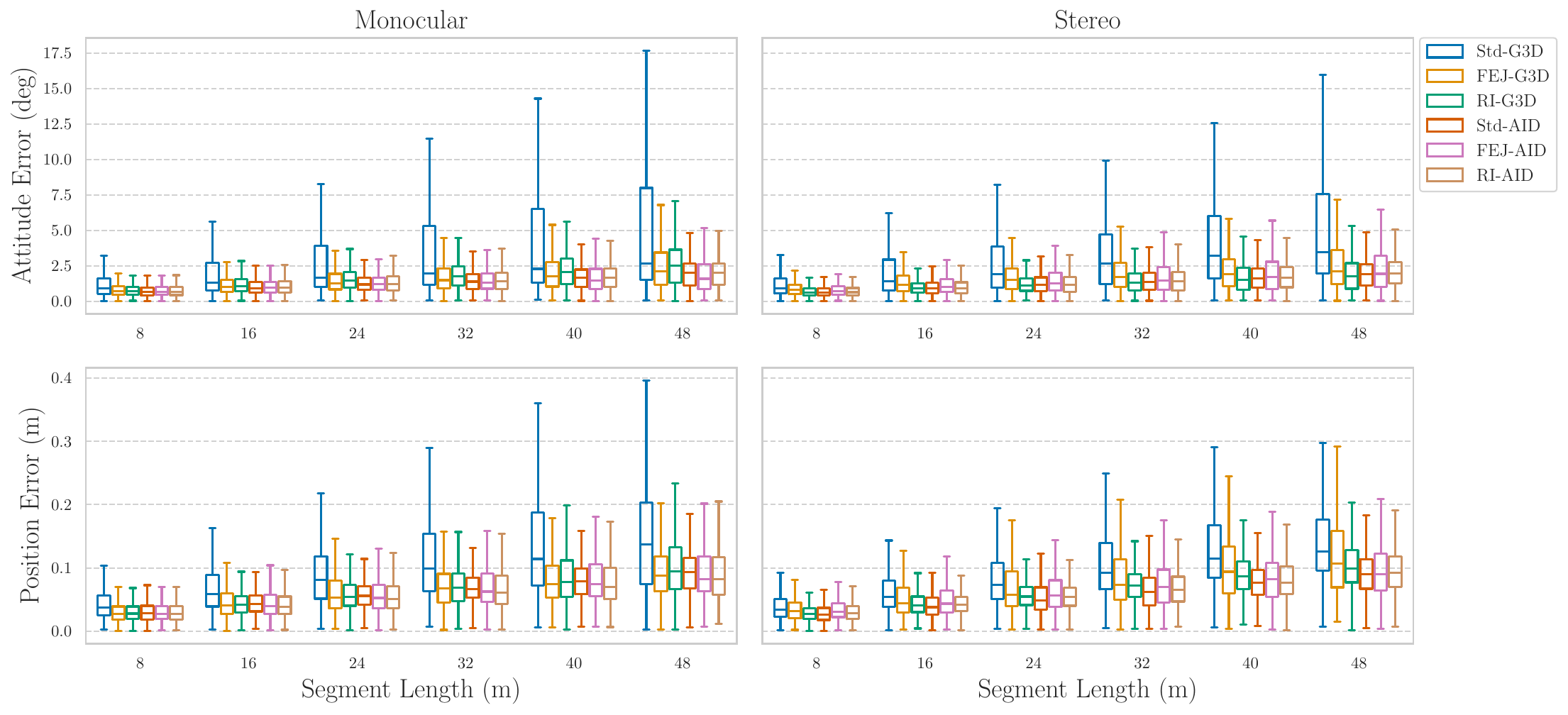}
    \caption{Distribution of RPEs for various trajectory segment lengths for each estimator configuration in both monocular (left) and stereo (right) configurations.}
    \label{fig:rpe_boxplot_tumvi}
\end{figure*}

Each estimator configuration is further evaluated on the \textbf{room} sequences of the TUM-VI dataset~\cite{schubert2018tum}, which provides hardware-synchronized monochrome fisheye stereo image pairs at $512 \times 512$ resolution and \SI{20}{\hertz} and IMU measurements at \SI{200}{\hertz} from a BMI160 IMU, with groundtruth poses from a motion capture system. The sliding window size is set to $c = 11$, and a maximum of 50 persistent SLAM features are kept in the state vector. Additionally, online calibration of the camera-IMU extrinsics and camera intrinsics is enabled. Each estimator is tested in both monocular and stereo configurations, where the default configuration for front-end feature tracking from~\cite{Geneva_Eckenhoff_Lee_Yang_Huang_2020} is used. The average ATE over 5 runs of each algorithm is shown in Table~\ref{tab:tum_vi_ate}. Additionally, to provide further insights into the performance of each estimator, the Relative Pose Error (RPE)~\cite{zhang2018tutorial} is computed for various segments of the trajectory length. The orientation and position RPE for both monocular and stereo configurations are shown in Fig.~\ref{fig:rpe_boxplot_tumvi}.

\begin{table*}[t!]
    \centering
    \caption{Average ATEs computed over 5 runs of each estimator, reported as orientation ATE (degrees) / position ATE (meters). The best result for each sequence and estimator is shown in \textbf{bold}, for both stereo and monocular configurations.}
    \label{tab:tum_vi_ate}
    \resizebox{0.7\textwidth}{!}{%
        \begin{tabular}{lcccccc}
            \toprule
            Algorithm & room1                  & room2                           & room3                           & room4                           & room5                  & Average                         \\
            \midrule
            \multicolumn{7}{c}{\textbf{Monocular}}                                                                                                                                                              \\
            \midrule
            Std-G3D   & \textbf{0.959} / 0.050 & 11.342 / 0.206                  & 8.230 / 0.160                   & 1.315 / 0.060                   & 1.612 / 0.094          & 4.692 / 0.114                   \\
            FEJ-G3D   & 1.325 / 0.056          & 3.882 / 0.094                   & \textbf{1.304} / 0.077          & 1.032 / 0.044                   & 1.605 / \textbf{0.069} & 1.830 / 0.068                   \\
            RI-G3D    & 1.506 / 0.062          & 3.643 / 0.103                   & 1.734 / 0.084                   & \textbf{0.923} / \textbf{0.030} & 1.877 / 0.091          & 1.937 / 0.074                   \\
            Std-AID   & 1.362 / 0.062          & \textbf{2.057} / \textbf{0.075} & 1.589 / \textbf{0.074}          & 1.022 / 0.037                   & 1.691 / 0.086          & 1.544 / 0.067                   \\
            FEJ-AID   & 1.297 / \textbf{0.048} & 2.828 / 0.097                   & 1.818 / 0.078                   & 0.935 / 0.040                   & \textbf{1.079} / 0.074 & 1.591 / 0.067                   \\
            RI-AID    & 1.128 / 0.049          & 2.530 / 0.080                   & 1.475 / 0.085                   & 1.017 / 0.034                   & 1.502 / 0.080          & \textbf{1.530} / \textbf{0.066} \\
            \midrule
            \multicolumn{7}{c}{\textbf{Stereo}}                                                                                                                                                                 \\
            \midrule
            Std-G3D   & 1.996 / 0.071          & 13.624 / 0.255                  & 5.030 / 0.122                   & 1.787 / 0.045                   & 1.991 / 0.099          & 4.886 / 0.118                   \\
            FEJ-G3D   & 2.001 / \textbf{0.056} & 5.032 / 0.181                   & 1.453 / 0.088                   & 1.169 / 0.040                   & \textbf{1.206} / 0.105 & 2.172 / 0.094                   \\
            RI-G3D    & \textbf{1.272} / 0.070 & 2.407 / \textbf{0.085}          & 1.397 / 0.086                   & 0.916 / 0.043                   & 1.468 / 0.113          & \textbf{1.492} / 0.079          \\
            Std-AID   & 1.374 / 0.062          & \textbf{2.310} / 0.117          & 1.477 / 0.085                   & 0.907 / 0.037                   & 1.542 / 0.082          & 1.522 / 0.076                   \\
            FEJ-AID   & 1.381 / 0.069          & 4.473 / 0.131                   & 1.565 / 0.079                   & \textbf{0.846} / \textbf{0.035} & 1.609 / \textbf{0.081} & 1.975 / 0.079                   \\
            RI-AID    & 1.727 / 0.068          & 2.792 / 0.092                   & \textbf{1.364} / \textbf{0.076} & 0.910 / 0.037                   & 1.612 / 0.091          & 1.681 / \textbf{0.073}          \\
            \bottomrule
        \end{tabular}
    }
\end{table*}

The results showcase that the Std-G3D configuration performs the worst on average, as changes in the landmark linearization points cause information gain along unobservable directions, ultimately leading to degraded performance. Specifically, Fig.~\ref{fig:rpe_boxplot_tumvi} shows that for both monocular and stereo configurations, the RPE for Std-G3D greatly grows, especially in orientation, for increased trajectory segment lengths. While the consistency-improved estimators FEJ-G3D and RI-G3D perform better than Std-G3D, their performance is still impacted by the use of FEJ on the landmark states, causing the Jacobians to be more inaccurate when the initial landmark estimate is poor. In contrast, it can be seen that all three estimators that utilize an anchored landmark parameterization perform similarly, including the standard estimator without any consistency improvements. Notably, the Std-AID estimator without any consistency improvements is able to obtain comparable or better performance than the consistency-improved estimators that use global feature representations. The results highlight that \textbf{the gap between estimator configurations utilizing anchored feature representations is significantly smaller than the gap between estimators employing a global feature representation}, due to the fact that the Std-AID estimator benefits from having a nullspace that is independent of the landmark states.
\section{Conclusion}
\label{sec:conclusion}
This paper analyzes the observability and consistency properties of EKF-based VINS estimators using anchored feature representations. The presented observability analysis reveals that the unobservable subspace of VINS estimators with anchored features does not depend on the landmarks, which leads to desirable consistency properties without requiring additional modifications. Two methods of further improving consistency for VINS estimators with anchored feature representations are presented, namely FEJ and the RI-EKF. Monte-Carlo simulations and real-world experiments on the TUM-VI dataset are used to compare the consistency-improved estimators that employ anchored feature representations against estimators that use global feature representations. Utilizing anchored feature representations is shown to offer consistency benefits, especially under large measurement noise. Future work will analyze the consistency properties of optimization-based estimators using anchored feature representations, in line with~\cite{chen2023optimization}.

% \clearpage
\appendix
% \vspace{-200mm}
\section{Nullspace of Observability Matrix}
\label{app:appendix}
To showcase the structure of the nullspace $\mbf{N}$, the structure of the observability matrix must be derived. The state vector for observability analysis is given by~\eqref{eq:analysis_state}, with $\mbs{\lambda}^1 = \mbf{r}_{c_f}^{p c_f}$, but the analysis can be extended to any anchored representation by considering different forms of $\mbf{h}_a \left(\mbf{r}_{c_f}^{p_1 c_f} \right)$. The discrete-time state transition matrix $\mbs{\Phi}_{k, 1}$ has the form
\begin{align}
    \mbs{\Phi}_{k, 1}        & =
    \begin{bmatrix}
        \mbs{\Phi}_{k, 1}^\mc{I} & \mbf{0}_{15 \times 9} \\
        \mbf{0}_{9 \times 15}    & \mbf{1}_{9 \times 9}
    \end{bmatrix},   \\
    \mbs{\Phi}_{k, 1}^\mc{I} & =
    \begin{bmatrix}
        \mbs{\Phi}^{11}_{k, 1} & \mbf{0}            & \mbf{0} & \mbs{\Phi}^{14}_{k, 1} & \mbf{0}                \\
        \mbs{\Phi}^{21}_{k, 1} & \mbf{1}            & \mbf{0} & \mbs{\Phi}^{24}_{k, 1} & \mbs{\Phi}^{25}_{k, 1} \\
        \mbs{\Phi}^{31}_{k, 1} & \Delta t_k \mbf{1} & \mbf{1} & \mbs{\Phi}^{34}_{k, 1} & \mbs{\Phi}^{35}_{k, 1} \\
        \mbf{0}                & \mbf{0}            & \mbf{0} & \mbf{1}                & \mbf{0}                \\
        \mbf{0}                & \mbf{0}            & \mbf{0} & \mbf{0}                & \mbf{1}
    \end{bmatrix},
\end{align}
where $\mbs{\Phi}_{k, 1}^\mc{I}$ is the state transition matrix for the inertial state, whose relevant blocks are given by
\begin{align}
    \mbs{\Phi}_{k, 1}^{11} & = \exp \left(\int_{t_1}^{t_k} \mbshat{\omega} \left(\tau \right) \mathrm{d} \tau \right) = \mbf{C}_{b_k b_1},                                                  \\
    \mbs{\Phi}_{k, 1}^{14} & = -\int_{t_1}^{t_k} \mbf{C}_{b_k b_\tau} \mathrm{d} \tau,                                                                                                      \\
    \mbs{\Phi}_{k, 1}^{31} & = -\left(\mbf{r}_a^{z_k w} - \mbf{r}_a^{z_1 w} - \mbf{v}_a^{z_k w/a} \Delta t_k - \frac{1}{2} \mbf{g}_a \Delta t_k^2 \right)^\times \mbf{C}_{ab_k},            \\
    \mbs{\Phi}_{k, 1}^{34} & = \int_{t_1}^{t_k} \int_{t_1}^\theta \mbf{C}_{ab_\tau} \mbfhat{a}_b^\times \int_{t_1}^{s} \mbf{C}_{b_s b_\tau} \mathrm{d} \tau \mathrm{d} s \mathrm{d} \theta, \\
    \mbs{\Phi}_{k, 1}^{35} & = \int_{t_1}^{t_k} \int_{t_1}^s \mbf{C}_{ab_\tau} \mathrm{d} \tau \mathrm{d} s,
\end{align}
and the full expressions for other block components can be found in~\cite{chen2022fej2}. Note that the analysis can also be performed by explicitly modeling the clone constraints in the state transition matrix as done in~\cite{lee2020visual}, and the conclusions of the analysis remain unchanged. The measurement Jacobian has the form
\begin{equation*}
    \small
    \mbf{H}_k = \mbf{H}^\pi \mbf{C}_{ab_k}^\trans
    \begin{bmatrix}
        \mbf{H}_k^\phi & \mbf{0} & -\mbf{1} & \mbf{0} & -\mbf{C}_{ab_f} \mbf{r}_b^{pz^\times} & \mbf{1} & \mbf{C}_{ab_f} \mbf{C}_{bc}
    \end{bmatrix},
\end{equation*}
where $\mbf{H}_k^\phi = \left(\mbf{r}_a^{p w} - \mbfbar{r}_a^{z_k w}\right)^\times \mbf{C}_{ab_k}$. With these forms of the discrete-time state transition matrix and measurement Jacobian, the $k$'th block row of the observability matrix is written as
\begin{equation*}
    \resizebox{\columnwidth}{!}{$
            \mbf{M}_k = \mbf{H}_k \mbs{\Phi}_{k, 1} = \mbs{\Gamma}_1
            \begin{bmatrix}
                \mbs{\Gamma}_2 & -\mbf{1} \Delta t & -\mbf{1} & \mbs{\Gamma}_3 & \mbs{\Gamma}_4 & -\mbf{C}_{ab_f} \mbf{r}_b^{pz^\times} & \mbf{1} & \mbf{C}_{ab_f} \mbf{C}_{bc}
            \end{bmatrix},
        $}
\end{equation*}
with the expressions for $\mbs{\Gamma}_1$, $\mbs{\Gamma}_2$, $\mbs{\Gamma}_3$, and $\mbs{\Gamma}_4$ given by
\begin{align}
    \mbs{\Gamma}_1 & = \mbf{H}^\pi \mbf{C}_{bc}^\trans \mbf{C}_{ab_k}^\trans,                                                                                 \\
    \mbs{\Gamma}_2 & = \mbf{H}^{\phi} \mbs{\Phi}_{k, 1}^{11} - \mbs{\Phi}_{k, 1}^{31},                                                                        \\
                   & = \left(\mbf{r}_a^{pw} - \mbf{r}_a^{z_1 w} - \mbf{v}_a^{z_1 w/a} \Delta t_k - \frac{1}{2} \mbf{g}_a \Delta t_k^2 \right) \mbf{C}_{ab_1}, \\
    \mbs{\Gamma}_3 & = \left(\mbf{r}_a^{pw} - \mbf{r}_a^{z_k w} \right)^\times \mbf{C}_{ab_k} \mbs{\Phi}_{k, 1}^{14} - \mbs{\Phi}_{k, 1}^{34}                 \\
    \mbs{\Gamma}_4 & = -\mbs{\Phi}_{k, 1}^{35}.
\end{align}
To verify that~\eqref{eq:ideal_nullspace} lies in the nullspace of the observability matrix, with $\mbf{N}_1^\lambda = \mbf{0}$ and $\mbf{N}_2^\lambda = \mbf{0}$, it can be verified that each block row of the observability matrix satisfies $\mbf{M}_k \mbf{N} = \mbf{0}$. Specifically, $\mbf{M}_k \mbf{N}^t = \mbs{\Gamma} \left(-\mbf{1} + \mbf{1} \right) = \mbf{0}$,
and
\begin{align}
    \begin{split}
        \mbf{M}_k \mbf{N}^r &= \mbs{\Gamma}_1 \left( \mbs{\Gamma}_2 \mbf{C}_{ab_1}^\trans \mbf{g}_a + \mbf{v}_a^{z_1 w/a^\times} \mbf{g}_a \Delta t_k \right. \\
        &\quad + \mbf{r}_a^{z_1 w^\times} \mbf{g}_a - \mbf{C}_{ab_f} \mbf{r}_b^{pz^\times} \mbf{C}_{ab_f}^\trans \mbf{g}_a \left. - \mbf{r}_a^{z_f w^\times}\mbf{g}_a \right),
    \end{split}                                                                                                                                                      \\
     & = \mbs{\Gamma}_1 \left(\mbf{r}_a^{pw^\times} \mbf{g}_a - \mbf{C}_{ab_f} \mbf{r}_b^{pz^\times} \mbf{C}_{ab_f}^\trans \mbf{g}_a - \mbf{r}_a^{z_f w^\times} \mbf{g}_a \right), \\
     & =\mbs{\Gamma}_1  \left(\mbf{r}_a^{pw^\times} \mbf{g}_a - \left(\mbf{C}_{ab_f} \mbf{r}_b^{pz} \right)^\times \mbf{g}_a - \mbf{r}_a^{z_f w^\times} \mbf{g}_a \right),         \\
     & = \mbs{\Gamma}_1 \left( \left(\mbf{r}_a^{pw} - \mbf{C}_{ab_f} \mbf{r}_b^{pz} - \mbf{r}_a^{z_f w} \right) \mbf{g}_a \right),                                                 \\
     & = \mbf{0}.
\end{align}
Since $\mbf{M}_k \mbf{N} = \mbf{0}, \forall k > 1$, $\mbf{N}$ belongs to the nullspace of the observability matrix $\mbf{M}$.

\printbibliography

@inproceedings{chen2025iteration,
  title        = {Is Iteration Worth It? {R}evisiting its Impact in Sliding-window {VIO}},
  author       = {Chen, Chuchu and Peng, Yuxiang and Huang, Guoquan},
  booktitle    = {Proc. IEEE Int. Conf. Robot. Automat.},
  pages        = {1060--1066},
  year         = {2025},
}

@article{qin2018vins,
  title     = {{VINS-Mono}: A Robust and Versatile Monocular Visual-Inertial State Estimator},
  author    = {Qin, Tong and Li, Peiliang and Shen, Shaojie},
  journal   = {IEEE Trans. Robot.},
  volume    = {34},
  number    = {4},
  pages     = {1004--1020},
  year      = {2018},
  publisher = {IEEE}
}

@article{Yang_Chen_Lee_Huang_2022,
  title        = {Decoupled Right-Invariant Error States for Consistent Visual-Inertial Navigation},
  volume       = {7},
  doi          = {10.1109/LRA.2021.3140054},
  abstractnote = {The invariant extended {Kalman} filter (IEKF) is proven to preserve the observability property of visual-inertial navigation systems (VINS) and suitable for consistent estimator design. However, if features are maintained in the state vector, the propagation of IEKF will become more computationally expensive because these features are involved in the covariance propagation. To address this issue, we propose two novel algorithms which preserve the system consistency by leveraging the invariant state representation and ensure efficiency by decoupling features from covariance propagation. The first algorithm combines right invariant error states with first-estimates Jacobian (FEJ) technique, by decoupling the features from the Lie group representation and utilizing FEJ for consistent estimation. The second algorithm is designed specifically for sliding-window filter-based VINS as it associates the features to an active cloned pose, instead of the current IMU state, for Lie group representation. A new pseudo-anchor change algorithm is also proposed to maintain the features in the state vector longer than the window span. Both decoupled right- and left-invariant error based VINS methods are implemented for a complete comparison. Extensive Monte-Carlo simulations on three simulated trajectories and real world evaluations on the TUM-VI datasets are provided to verify our analysis and demonstrate that the proposed algorithms can achieve improved accuracy than a state-of-art filter-based VINS algorithm using FEJ.},
  number       = {2},
  journal      = {IEEE Robot. Automat. Lett.},
  author       = {Yang, Yulin and Chen, Chuchu and Lee, Woosik and Huang, Guoquan},
  year         = {2022},
  month        = apr,
  pages        = {1627–1634}
}

@article{Civera_Davison_Montiel_2008,
  title        = {Inverse Depth Parametrization for Monocular SLAM},
  volume       = {24},
  doi          = {10.1109/TRO.2008.2003276},
  abstractnote = {We present a new parametrization for point features within monocular simultaneous localization and mapping (SLAM) that permits efficient and accurate representation of uncertainty during undelayed initialization and beyond, all within the standard extended Kalman filter (EKF). The key concept is direct parametrization of the inverse depth of features relative to the camera locations from which they were first viewed, which produces measurement equations with a high degree of linearity. Importantly, our parametrization can cope with features over a huge range of depths, even those that are so far from the camera that they present little parallax during motion—maintaining sufficient representative uncertainty that these points retain the opportunity to "come in” smoothly from infinity if the camera makes larger movements. Feature initialization is undelayed in the sense that even distant features are immediately used to improve camera motion estimates, acting initially as bearing references but not permanently labeled as such. The inverse depth parametrization remains well behaved for features at all stages of SLAM processing, but has the drawback in computational terms that each point is represented by a 6-D state vector as opposed to the standard three of a Euclidean XYZ representation. We show that once the depth estimate of a feature is sufficiently accurate, its representation can safely be converted to the Euclidean XYZ form, and propose a linearity index that allows automatic detection and conversion to maintain maximum efficiency—only low parallax features need be maintained in inverse depth form for long periods. We present a real-time implementation at 30 Hz, where the parametrization is validated in a fully automatic 3-D SLAM system featuring a handheld single camera with no additional sensing. Experiments show robust operation in challenging indoor and outdoor},
  number       = {5},
  journal      = {IEEE Trans. Robot.},
  author       = {Civera, Javier and Davison, Andrew J. and Montiel, J. M. MartÍnez},
  year         = {2008},
  month        = oct,
  pages        = {932–945}
}

@inproceedings{Geneva_Eckenhoff_Lee_Yang_Huang_2020,
  title        = {{OpenVINS}: A Research Platform for Visual-Inertial Estimation},
  url          = {https://ieeexplore.ieee.org/document/9196524/},
  doi          = {10.1109/ICRA40945.2020.9196524},
  abstractnote = {In this paper, we present an open platform, termed OpenVINS, for visual-inertial estimation research for both the academic community and practitioners from industry. The open sourced codebase provides a foundation for researchers and engineers to quickly start developing new capabilities for their visual-inertial systems. This codebase has out of the box support for commonly desired visual-inertial estimation features, which include: (i) on-manifold sliding window Kalman filter, (ii) online camera intrinsic and extrinsic calibration, (iii) camera to inertial sensor time offset calibration, (iv) SLAM landmarks with different representations and consistent First-Estimates Jacobian (FEJ) treatments, (v) modular type system for state management, (vi) extendable visual-inertial system simulator, and (vii) extensive toolbox for algorithm evaluation. Moreover, we have also focused on detailed documentation and theoretical derivations to support rapid development and research, which are greatly lacked in the current open sourced algorithms. Finally, we perform comprehensive validation of the proposed OpenVINS against state-of-the-art open sourced algorithms, showing its competing estimation performance.},
  booktitle    = {Proc. IEEE Int. Conf. Robot. Automat.},
  author       = {Geneva, Patrick and Eckenhoff, Kevin and Lee, Woosik and Yang, Yulin and Huang, Guoquan},
  year         = {2020},
  month        = may,
  pages        = {4666–4672}
}

@inproceedings{Mourikis_Roumeliotis_2007,
  title        = {A Multi-State Constraint {Kalman} Filter for Vision-aided Inertial Navigation},
  url          = {https://ieeexplore.ieee.org/document/4209642/},
  doi          = {10.1109/ROBOT.2007.364024},
  abstractnote = {In this paper, we present an extended Kalman filter (EKF)-based algorithm for real-time vision-aided inertial navigation. The primary contribution of this work is the derivation of a measurement model that is able to express the geometric constraints that arise when a static feature is observed from multiple camera poses. This measurement model does not require including the 3D feature position in the state vector of the EKF and is optimal, up to linearization errors. The vision-aided inertial navigation algorithm we propose has computational complexity only linear in the number of features, and is capable of high-precision pose estimation in large-scale real-world environments. The performance of the algorithm is demonstrated in extensive experimental results, involving a camera/IMU system localizing within an urban area.},
  booktitle    = {Proc. IEEE Int. Conf. Robot. Automat.},
  author       = {Mourikis, Anastasios I. and Roumeliotis, Stergios I.},
  year         = {2007},
  month        = apr,
  pages        = {3565–3572}
}

@article{Leutenegger_Lynen_Bosse_Siegwart_Furgale_2015,
  title        = {Keyframe-based visual–inertial odometry using nonlinear optimization},
  volume       = {34},
  doi          = {10.1177/0278364914554813},
  abstractnote = {Combining visual and inertial measurements has become popular in mobile robotics, since the two sensing modalities offer complementary characteristics that make them the ideal choice for accurate visual–inertial odometry or simultaneous localization and mapping (SLAM). While historically the problem has been addressed with filtering, advancements in visual estimation suggest that nonlinear optimization offers superior accuracy, while still tractable in complexity thanks to the sparsity of the underlying problem. Taking inspiration from these findings, we formulate a rigorously probabilistic cost function that combines reprojection errors of landmarks and inertial terms. The problem is kept tractable and thus ensuring real-time operation by limiting the optimization to a bounded window of keyframes through marginalization. Keyframes may be spaced in time by arbitrary intervals, while still related by linearized inertial terms. We present evaluation results on complementary datasets recorded with our custom-built stereo visual–inertial hardware that accurately synchronizes accelerometer and gyroscope measurements with imagery. A comparison of both a stereo and monocular version of our algorithm with and without online extrinsics estimation is shown with respect to ground truth. Furthermore, we compare the performance to an implementation of a state-of-the-art stochastic cloning sliding-window filter. This competitive reference implementation performs tightly coupled filtering-based visual–inertial odometry. While our approach declaredly demands more computation, we show its superior performance in terms of accuracy.},
  number       = {3},
  journal      = {Int. J. Robot. Res.},
  author       = {Leutenegger, Stefan and Lynen, Simon and Bosse, Michael and Siegwart, Roland and Furgale, Paul},
  year         = {2015},
  month        = mar,
  pages        = {314–334},
}

@inproceedings{huang2009first,
  title        = {A First-Estimates {J}acobian {EKF} for Improving SLAM Consistency},
  author       = {Huang, Guoquan P and Mourikis, Anastasios I and Roumeliotis, Stergios I},
  booktitle    = {Proc. Int. Symp. Exp. Robot.},
  pages        = {373--382},
  year         = {2009},
  organization = {Springer}
}

@inproceedings{Huang_Mourikis_Roumeliotis_2011,
  title        = {An Observability-Constrained Sliding Window Filter for {SLAM}},
  url          = {https://ieeexplore.ieee.org/document/6095161/},
  doi          = {10.1109/IROS.2011.6095161},
  abstractnote = {A sliding window filter (SWF) is an appealing smoothing algorithm for nonlinear estimation problems such as simultaneous localization and mapping (SLAM), since it is resource-adaptive by controlling the size of the sliding window, and can better address the nonlinearity of the problem by relinearizing available measurements. However, due to the marginalization employed to discard old states from the sliding window, the standard SWF has different parameter observability properties from the optimal batch maximum-a-posterior (MAP) estimator. Specifically, the nullspace of the Fisher information matrix (or Hessian) has lower dimension than that of the batch MAP estimator. This implies that the standard SWF acquires spurious information, which can lead to inconsistency. To address this problem, we propose an observability-constrained (OC)-SWF where the linearization points are selected so as to ensure the correct dimension of the nullspace of the Hessian, as well as minimize the linearization errors. We present both Monte Carlo simulations and real-world experimental results which show that the OC-SWF’s performance is superior to the standard SWF, in terms of both accuracy and consistency.},
  booktitle    = {Proc. IEEE/RSJ Int. Conf. Intell. Robot. Syst.},
  author       = {Huang, Guoquan P. and Mourikis, Anastasios I. and Roumeliotis, Stergios I.},
  year         = {2011},
  month        = sept,
  pages        = {65–72}
}

@article{Solà_Deray_Atchuthan_2021,
  title        = {A Micro {L}ie Theory for State Estimation in Robotics},
  url          = {http://arxiv.org/abs/1812.01537},
  doi          = {10.48550/arXiv.1812.01537},
  abstractnote = {A Lie group is an old mathematical abstract object dating back to the XIX century, when mathematician Sophus Lie laid the foundations of the theory of continuous transformation groups. Its inﬂuence has spread over diverse areas of science and technology many years later. In robotics, we are recently experiencing an important trend in its usage, at least in the ﬁelds of estimation, and particularly in motion estimation for navigation. Yet for a vast majority of roboticians, Lie groups are highly abstract constructions and therefore difﬁcult to understand and to use.},
  note         = {arXiv:1812.01537},
  publisher    = {arXiv},
  author       = {Solà, Joan and Deray, Jeremie and Atchuthan, Dinesh},
  year         = {2021}
}

@article{prentice2009belief,
  title     = {The Belief Roadmap: Efficient Planning in Belief Space by Factoring the Covariance},
  author    = {Prentice, Samuel and Roy, Nicholas},
  journal   = {Int. J. Robot. Res.},
  volume    = {28},
  number    = {11-12},
  pages     = {1448--1465},
  year      = {2009},
  publisher = {Sage Publications Sage UK: London, England}
}

@inproceedings{doherty2019multimodal,
  title={Multimodal Semantic {SLAM} with Probabilistic Data Association},
  author={Doherty, Kevin and Fourie, Dehann and Leonard, John},
  booktitle={Proc. IEEE Int. Conf. Robot. Automat.},
  pages={2419--2425},
  year={2019},
}

@inproceedings{paul2017comparative,
  title={A Comparative Analysis of Tightly-Coupled Monocular, Binocular, and Stereo {VINS}},
  author={Paul, Mrinal K and Wu, Kejian and Hesch, Joel A and Nerurkar, Esha D and Roumeliotis, Stergios I},
  booktitle={Proc. IEEE Int. Conf. Robot. Automat.},
  pages={165--172},
  year={2017},
}

@inproceedings{chen1990local,
  title={Local Observability matrix and its Application to Observability Analyses},
  author={Chen, Zhe and Jiang, Ke and Hung, James C},
  booktitle={Proc. IEEE Int. Conf. Ind. Electron. Soc.},
  pages={100--103},
  year={1990},
}

@inproceedings{schubert2018tum,
  title={The {TUM-VI} Benchmark for Evaluating Visual-Inertial Odometry},
  author={Schubert, David and Goll, Thore and Demmel, Nikolaus and Usenko, Vladyslav and St{\"u}ckler, J{\"o}rg and Cremers, Daniel},
  booktitle={Proc. IEEE/RSJ Int. Conf. Intell Robot. Syst.},
  pages={1680--1687},
  year={2018},
}

@article{huai2022robocentric,
  title={Robocentric Visual--Inertial Odometry},
  author={Huai, Zheng and Huang, Guoquan},
  journal={Int. J. Robot. Res. (IJRR)},
  volume={41},
  number={7},
  pages={667--689},
  year={2022},
  publisher={SAGE Publications Sage UK: London, England}
}

@inproceedings{chen2022fej2,
  title={{FEJ2}: A Consistent Visual-Inertial State Estimator Design},
  author={Chen, Chuchu and Yang, Yulin and Geneva, Patrick and Huang, Guoquan},
  booktitle={Proc. IEEE Int. Conf. Robot. Automat.},
  pages={9506--9512},
  year={2022},
}

@article{hesch2013consistency,
  title={Consistency Analysis and Improvement of Vision-Aided Inertial Navigation},
  author={Hesch, Joel A and Kottas, Dimitrios G and Bowman, Sean L and Roumeliotis, Stergios I},
  journal={IEEE Trans. Robot.},
  volume={30},
  number={1},
  pages={158--176},
  year={2013},
}

@article{li2013high,
  title={{High-precision, consistent EKF-based visual-inertial odometry}},
  author={Li, Mingyang and Mourikis, Anastasios I},
  journal={Int. J. Robot. Res.},
  volume={32},
  number={6},
  pages={690--711},
  year={2013},
  publisher={Sage Publications Sage UK: London, England}
}

@inproceedings{roumeliotis2002stochastic,
  title={Stochastic Cloning: A Generalized Framework for Processing Relative State Measurements},
  author={Roumeliotis, Stergios I and Burdick, Joel W},
  booktitle={Proc. IEEE Int. Conf. Robot. Automat.},
  volume={2},
  pages={1788--1795},
  year={2002},
}

@phdthesis{li2014visual,
  title={Visual-Inertial Odometry on Resource-Constrained Systems},
  author={Li, Mingyang},
  year={2014},
  publisher={University of California, Riverside}
}

@inproceedings{zhang2018tutorial,
  title={A Tutorial on Quantitative Trajectory Evaluation for Visual(-Inertial) Odometry},
  author={Zhang, Zichao and Scaramuzza, Davide},
  booktitle={Proc. IEEE/RSJ Int. Conf. Intell Robot. Syst.},
  pages={7244--7251},
  year={2018},
}

@article{sola2012impact,
  title={Impact of Landmark Parameterization on Monocular {EKF-SLAM} with Points and Lines},
  author={Sola, Joan and Vidal-Calleja, Teresa and Civera, Javier and Montiel, Jos{\'e} Mar{\'\i}a Mart{\'\i}nez},
  journal={Int. J. Comput. Vis.},
  volume={97},
  number={3},
  pages={339--368},
  year={2012},
  publisher={Springer}
}

@inproceedings{montiel2006unified,
  title={Unified Inverse Depth Parametrization for Monocular {SLAM}},
  author={Montiel, JM Mart{\'\i}nez and Civera, Javier and Davison, Andrew J},
  booktitle={Robotics: Science and Systems},
  volume={3},
  pages={1--8},
  year={2006}
}

@inproceedings{wu2017invariant,
  title={{An invariant-EKF VINS algorithm for improving consistency}},
  author={Wu, Kanzhi and Zhang, Teng and Su, Daobilige and Huang, Shoudong and Dissanayake, Gamini},
  booktitle={Proc. IEEE/RSJ Int. Conf. Intell Robot. Syst.},
  pages={1578--1585},
  year={2017},
}

@book{farrell2008aided,
  title={Aided Navigation: GPS with High-Rate Sensors},
  author={Farrell, Jay},
  year={2008},
  publisher={McGraw-Hill, Inc.}
}

@inproceedings{chen2023optimization,
  title={{Optimization-Based {VINS}: consistency, marginalization, and {FEJ}}},
  author={Chen, Chuchu and Geneva, Patrick and Peng, Yuxiang and Lee, Woosik and Huang, Guoquan},
  booktitle={Proc. IEEE/RSJ Int. Conf. Intell. Robot. Syst.},
  pages={1517--1524},
  year={2023},
}

@Article{peng2025sqrt,
  title={{$\sqrt{\mathrm{VINS}}$}: Robust and Ultrafast Square-Root Filter-Based 3D Motion Tracking},
  author    = {Peng, Yuxiang and Chen, Chuchu and Wu, Kejian and Huang, Guoquan},
  journal   = {IEEE Trans. Robot.},
  year      = {2025},
  volume = {41},
  pages = {6570--6589},
}

@article{tian2025t,
  title={{T-ESKF: transformed error-state {Kalman} filter for consistent visual-inertial navigation}},
  author={Tian, Chungeng and Hao, Ning and He, Fenghua},
  journal={IEEE Robot. Automat. Lett.},
  volume={10},
  number={2},
  pages={1808--1815},
  year={2025},
}

@article{song2026affine,
  title={{Affine EKF: Exploring and utilizing sufficient and necessary conditions for observability maintenance to improve EKF consistency}},
  author={Song, Yang and Zhao, Liang and Huang, Shoudong},
  journal={IEEE Trans. Robot.},
  year={2026},
}

@inproceedings{lee2020visual,
  title={Visual-inertial-wheel odometry with online calibration},
  author={Lee, Woosik and Eckenhoff, Kevin and Yang, Yulin and Geneva, Patrick and Huang, Guoquan},
  booktitle={IEEE/RSJ Int. Conf. Intell. Robot. Syst.},
  pages={4559--4566},
  year={2020},
}

\end{document}